\documentclass[11pt,onecolumn]{article}
\usepackage{booktabs}       
\usepackage{caption} 
\usepackage{footnote}
\makesavenoteenv{tabular}
\usepackage[utf8]{inputenc}
\usepackage{times}
\usepackage{amsmath}
\usepackage{amssymb}
\usepackage{multirow}
\usepackage{xspace}
\usepackage{theorem}
\usepackage{graphicx}
\usepackage{ifpdf}
\usepackage{url,hyperref}
\usepackage{latexsym}
\usepackage{euscript}
\usepackage{xspace}
\usepackage{color}
\usepackage{makeidx}
\usepackage{wrapfig}
\usepackage{xypic}
\xyoption{all}

\long\def\remove#1{}

\definecolor{darkred}{rgb}{1, 0.1, 0.3}
\definecolor{darkgreen}{rgb}{0.5, 0.8, 0.1}
\definecolor{darkpurple}{rgb}{1.0, 0, 1.0}
\definecolor{darkblue}{rgb}{0, 0, 1.0}

\newcommand {\mm}[1] {\ifmmode{#1}\else{\mbox{\(#1\)}}\fi}

\newcommand{\DOM}       {{LDP}\xspace}

\title{A simple yet effective baseline for non-attributed graph classification}
\author{Chen Cai \footnote{The shorter version of this paper appears on 2019 ICLR Workshop: Representation Learning on Graphs and Manifolds \url{https://rlgm.github.io/papers/}}\\
The Ohio State University\\
\texttt{cai.507@osu.edu} \\
\and
Yusu Wang  \\
The Ohio State University\\
\texttt{yusu@cse.ohio-state.edu} \\
}
\date{}

\begin{document}

\maketitle
\begin{abstract}
Graphs are complex objects that do not lend themselves easily to typical learning tasks. Recently, a range of approaches based on graph kernels or graph neural networks have been developed for graph classification and for representation learning on graphs in general. As the developed methodologies become more sophisticated, it is important to understand which components of the increasingly complex methods are necessary or most effective.

As a first step, we develop a simple yet meaningful graph representation, and explore its effectiveness in graph classification. We test our baseline representation for the graph classification task on a range of graph datasets.  Interestingly, this simple representation achieves similar performance as the state-of-the-art graph kernels and graph neural networks for non-attributed graph classification. Its performance on classifying attributed graphs is slightly weaker as it does not incorporate attributes. However, given its simplicity and efficiency, we believe that it still serves as an effective baseline for attributed graph classification. Our graph representation is efficient (linear-time) to compute. We also provide a simple connection with the graph neural networks.

Note that these observations are only for the task of graph classification while existing methods are often designed for a broader scope including node embedding and link prediction. The results are also likely biased due to the limited amount of benchmark datasets available.  Nevertheless, the good performance of our simple baseline calls for the development of new, more comprehensive benchmark datasets so as to better evaluate and analyze different graph learning methods. Furthermore, given the computational efficiency of our graph summary, we believe that it is a good candidate as a baseline method for future graph classification (or even other graph learning) studies.     
\end{abstract} 

\section{Introduction}
Graph-type data are ubiquitous across many scientific fields. For example, social networks, molecular graphs, biological protein-protein interaction networks, knowledge graphs, and recommender systems, are all graph objects naturally arise in a range of application domains, where edges can describe the interaction and relationship between individual entities. 

Recently, there has been a surge of approaches that aim to learn representations that encode structure information about the graph. On the high level, the methods can be categorized to be graph-kernel based, or graph neural network based. 
The problem remains challenging. For example, the design of graph kernel and graph neural networks are often influenced by the (sub-)graph isomorphism problem, where one aims to make sure that features of non-isomorphic graphs are likely different. On the one hand, graph isomorphism is computationally hard \footnote{The graph isomorphism testing itself has only very recently shown to be solvable in quasi-polynomial time \cite{babai2016graph}. The subgraph isomorphism, on the other hand, is NP-complete.}. On the other hand, there is no guarantee that graph kernels/neural networks with the strongest expressive power in terms of differentiating non-isomorphic graphs will generalize the best.
In general, one also has to strike a balance between the flexibility and (over-)expressive power of a graph representation framework. In addition, the method should also be computationally efficient, scalable to large datasets. Several existing graph kernels rely on certain spectral structures of the adjacency matrix of the graph, and can be computationally expensive. 

Nevertheless, much progress has been made in designing more sensitive graph kernels or more expressive graph neural networks. However, as the methodologies become more and more sophisticated, it becomes harder to understand which components of these methods are more crucial. For example, graph neural network (GNN) based approaches are flexible, and can scale to large datasets. They achieve state-of-the-art performance in several tasks including node classification, link prediction, and graph classification. However, the architecture of graph neural networks mixes representation and optimization, making it hard to analyze its power and limitation rigorously. 

In this paper, we are interested in evaluating existing approaches for graph classification tasks, in an attempt to gain a gradual understanding of their powers and limitations. In an effort to do so, we develop a simple graph representation based on local information for non-attributed graphs, which we refer to as \DOM{}. One of our initial goals was to understand in which scenarios the simple summary shall fail. 
To this end, we collect all the graph datasets from existing literature on graph classification; see Section \ref{sec: dataset} for the description of the datasets we use. 

Interestingly, this simple graph representation achieves similar performance as the state-of-the-art graph kernels and graph neural networks for non-attributed graph classification, and in fact, outperforms many existing more sophisticated representations. Its performance on classifying attributed graphs (mostly graphs representing biochemical compounds/molecules) is also competitive despite that it does not incorporate attributes. 
We report these results as we believe they make the following contributions: 
\begin{itemize}
    \item While our LDP graph representation is simple, it is intuitive and we show its competitive performance in graph classification for a range of graph datasets. This graph representation is computationally efficient (linear-time) to compute. We also provide a simple connection between our representation with the graph neural networks. 
    \item We note that these observations are only for the task of graph classification; while existing methods (especially graph neural network based approaches) are often designed for a broader scope, including for node embedding and for link prediction purposes. The results are also likely biased due to the limited amount of benchmark datasets available, and thus do not form the basis to dismiss any existing graph classification methodology.  

    Nevertheless, the good performance of our simple graph representation raises concerns about the effectiveness of current benchmark datasets for evaluating different algorithms for non-attribute graph classification. It calls for the development of new, more comprehensive, benchmark datasets for the better evaluation of different graph learning methods as well as for more rigorous analysis of the power and limitation of graph representations. 
    \item Furthermore, given the computational efficiency of our graph summary, we believe that it is a good candidate as a baseline method for future graph classification (or even other graph learning) studies.  
\end{itemize}

\section{Related Work} 
\subsection{Graph Kernel}
There are various graph kernels, many of which explore the R-convolutional framework\cite{haussler1999convolution}. The key idea is to decompose a whole graph into small substructures and build graph kernels based on the similarities defined for these components. Graph kernels following this line of work differ from each other in the way they decompose graphs. For example, graphlet kernels\cite{shervashidze2009efficient} are based on small subgraphs up to a fixed size. Weisfeiler-Lehman graph kernels\cite{shervashidze2011weisfeiler} is based on subtree patterns. Shortest path kernel\cite{borgwardt2005shortest} is derived by comparing the paths between graphs.  Other graph kernels, such as\cite{vishwanathan2010graph} and\cite{gartner2003graph}, are developed by counting the number of common random walks on direct product graphs. However, all the above R-convolution based graph kernels suffer from a drawback. As pointed out in\cite{yanardag2015structural}, increasing the size of substructures will largely decrease the probability that two graphs contain similar substructures, which usually results in the ``diagonal dominance issue'' \cite{kandola2003reducing}.  More recently, new methods have been proposed to compare graphs, which is done by quantifying the dissimilarity between the distributions of pairwise distances between nodes. \cite{schieber2017quantification} uses the shortest path distance, and\cite{verma2017hunt} uses the diffusion distance. In general, most graph kernels can handle label information, but there are a few exceptions\cite{shervashidze2009efficient, verma2017hunt} that purely use graph topology.

\subsection{Graph Neural Network}
Another way to tackle graph classification involves developing graph neural network (GNN). GNN broadly follows a recursive neighborhood aggregation(or message passing) scheme, where each node aggregates feature vectors of its neighbors to compute its new feature vector. Repeating the above procedure $k$ times, a node is represented by its transformed feature vector. The representation of an entire graph can be obtained by pooling, for example, by summing the feature vectors over all nodes in the graph. 

Many GNN variants with different neighborhood aggregation and graph-level pooling schemes have been proposed \cite{defferrard2016convolutional, duvenaud2015convolutional, hamilton2017inductive, li2015gated, kearnes2016molecular, kipf2016semi, velickovic2017graph, ying2018hierarchical}. 
Empirically, these GNNs have achieved state-of-the-art performance in many tasks such as node classification, link prediction, and graph classification. However, the design of new GNNs is often based on empirical intuition, heuristics, and experimental trial-and-error. The theoretical understanding of the properties and limitations of GNNs is somewhat limited, although there exists very recent work starting to address this issue \cite{xu2018powerful}. 

\subsection{Baseline method}
We remark that there has been a few works that draw attention to the potential problems of widely used graph datasets. Specifically, \cite{yuliia2015graph} observe that even by using simple graph features such as the number of nodes, they could achieve similar classification performance on common benchmark datasets compared to early graph kernels.  Their results show that we cannot solely rely on these data sets to show the performance of a graph kernel. However, their work focuses on attributed graphs, while our work focuses on non-attributed graphs. Very recently, in an independent work by \cite{wu2019simplifying}, the authors hypothesize that the nonlinearity between GCN layers is not critical. They remove the non-linearities and develop a simple network called Simple Graph Convolution. This network works well on text classification, semi-supervised user geolocation, relation extract, zero-shot image classification, and graph classification. While our work shares some similarities with theirs, our paper was developed independently. 

\section{A simple baseline for graph classification}
\subsection{Our model}
We denote a graph as $G(V, E)$ where $V$ is the set of nodes and $E$ is the set of edges in $G$. For each graph $G$ we extract features for each node in the following way. For each node $v \in G(V)$, let $DN(v)$ denote the multiset of the \textbf{d}egree of all the \textbf{n}eighboring nodes of $v$, i.e., $DN(v) = \{\text{degree}(u) | (u, v) \in E \}$. We take five node features, which are (degree($v$), min(DN($v$)), max(DN($v$)),mean(DN($v$)), std(DN($v$))). In other words, each node feature summarizes the degree information of this node and its 1- neighborhood. We aggregate features over different nodes in the same graph by performing either a histogram or an empirical distribution function(edf) operation, i.e, mapping all node feature into a histogram or an empirical distribution. 

We then repeat the same procedure for all five node features and concatenate all the feature vectors as the input feature for SVM classifier. For a fair comparison, we follow the convention in the graph kernel literature. We perform 10-fold cross validation ten times and report the average accuracy. For simplicity, in the rest of the paper, we denote our baseline as \textbf{ Local Degree Profile(LDP)}.

\textbf{Computational complexity}:
In feature extraction, we only need to count the degree for each node and save the statistics of 1- neighborhood for each node. This can be done in $O(E)$ time. To map $V$ numbers into $B$ bin takes $O(V)$ time so the total complexity is $O(E)$ time. This matches the lower bound of reading a graph. In comparison, we attach the table summarizing the complexity of computing various graph kernels, both exactly and approximately.
\footnote{Deep graph kernel requires corpus generation to build the kernel and we do not find explicit computational complexity in the original paper so it is not listed in the table. The kernels in the table are WL(Weisfeiler-Lehman kernel)\cite{shervashidze2011weisfeiler}, WL-OA (Weisfeiler-Lehman optimal assignment kernel)\cite{kriege2016valid}, GK(Graphlet Kernel)\cite{shervashidze2009efficient}, RetGK(Graph Kernels based on Return Probabilities
of Random Walks)\cite{zhang2018retgk}, MLG(Multiscale Laplacian Kernel)\cite{kondor2016multiscale}, FGSD(family of graph spectral distances)\cite{verma2017hunt}.} 

The more computationally expensive part of our algorithm is the SVM. For linear SVM it takes $O(nd^2)$ time where $n$ is the number of features and $d$ is the dimension of features. For non-linear kernel it is $O(n^2d)$. Although there is an efficient algorithm available to approximate the feature map\cite{rahimi2008random}, we can still afford running the original algorithm in a relatively short time. 
\begin{table}[h]
\label{tab: t1}
\centering
\begin{tabular}{@{}lllllllll@{}}
\toprule
Complexity  
& WL/WL-OA           
& GK                      
& RetGK      
& MLG                       
& FGSD    
& \DOM{} \\
Approximate & -                       & $O(Vd^{k-1}) $             & $O(D + d)$                       & $O(V^3)$                 & $O(rE)$                 & -        \\
Worst-Case  & $O(hE)$            & $O(V^k)$                        &  $O(V^2)$                          &  $O(\tilde{V}^3)$       & $O(V^3)$                  & $O(E)$   \\ \bottomrule
\end{tabular}
\caption{ Complexity of various graph kernels. With a slight abuse of notation, $V$ is the number of nodes and $E$ is the number of edges in the larger graph among two graphs. For WL\cite{shervashidze2011weisfeiler, kriege2016valid}, $h$ is the number of iterations. For GK\cite{shervashidze2009efficient}, $d < V$ and $k\in\{3,4,5\}$.  For RetGK\cite{zhang2018retgk}, return probabilities of random walks need to be calculated before computing the kernel, which takes $O(V^3 + (S + 1)V^2)$ exactly where $S$ is the number of steps, and takes $O(VSM)$ approximately where $M$ is number of Monte Carlo simulations used for simulation of random walks. $D$ is the number of random Fourier features and $d$ is the dimension of input feature. For MLG\cite{kondor2016multiscale}, $\tilde{V}$ is the number of sampled vertices and $\tilde{V} < V.$ For FGSD\cite{verma2017hunt}, $r$ the number of terms of polynomials used to approximate $f(L).$} 
\end{table}

\textbf{Hyperparameter: }
Below we describe the hyper-parameters of our model. They arise naturally when discretizing continuous node features on graphs for the down-stream classifier. In practice, they are robust and easy to tune. 

\emph{Bin size.}  In our experiment, we map the neighborhood degree distribution into a different number of $n$ bins of uniform width. We try different sizes from $\{30, 50, 70, 100\}$ to discretize the distribution.

\emph{Normalization.}  There are two natural ways of normalization for our method. The first one is to normalize every graph separately so that the value represents the relative degree. The second one is to normalize the whole dataset by finding the largest degree value across all the graphs. In practice, we do not see consistent advantage of one normalization over the other so we try both and pick the one gives best training accuracy.

\emph{Empirical Distribution versus Histogram.} We try to represent node features over a graph by both a histogram or an empirical distribution because empirical distribution is more stable with respect with the particular choice of bin size. It indeed yields better results than the histogram on certain dataset so in experiments we treat this choice as a hyperparameter. 

\emph{Linear vs logarithmic scales.} The degree distribution of many real-life networks follows the power law, which is usually visualized by log-log scale. We try both log scale and linear scale and notice that log scale yields better results for REDDIT 5K and REDDIT 12K. 

\emph{SVM parameter.} The $C$ parameter is selected from $\{10^{-3}, 10^{-2}, ..., 10^{2}, 10^{3}\}$ and the Gaussian bandwidth is selected from $\{10^{-2}, 10^{-1}, 1, 10^{1}, 10^{2}\}$. 

\emph{Remark.} In general, all those hyper-parameters except the SVM parameters are not sensitive.  Fine tuning of the above hyper-parameters usually yields about an improvement of 2 percentage.

\subsection{Relation to graph neural networks }
Our baseline can be seen as a variant of GNN that is used to learn useful representations for graph classification in an end-to-end manner.  GNNs use the graph structure $G$ and node features $X_{v}$ to learn either node-level representation or graph-level representation. Formally, the $k$-th layer of a GNN is
\begin{equation}
a_{v}^{(k)} = \text{AGGREGATE}^{(k)}(\{ h_{u}^{(k-1)} : u \in N(u) \}), h_{v}^{(k)} = \text{COMBINE}^{(k)}(h_{v}^{(k-1)}, a_v^{(k)})
\end{equation}
where $h_{v}^{(k)}$ is the feature vector of node $v$ at the $k$-th layer. $h_v^{(0)} = X_v$ initially, and $N(v)$ is neighbors of $v$. The choices of $\text{AGGREGATE}^{(k)}(\cdot) $ and $\text{COMBINE}^{(k)}(\cdot)$ in GNNs are crucial. A number of architectures for AGGREGATE have been proposed. For example, one popular variant of GNN Graph Convolutional Networks (GCNs) \cite{kipf2016semi}, implements AGGREGATE as
$a_v^{(k)} = \text{MEAN}(\{ \text{ReLU}(W \cdot h_u^{(k-1)}), \forall u \in N(v)  \})$
where $W$ is a learnable weight matrix. 

For node classification, the node representation $h_v^{(K)}$ of the last layer is used for prediction. For graph classification task, an additional READOUT function is needed to aggregate node features into a graph representation $h_G = \text{READOUT}(\{ h_v^{(K)} | v \in G  \})$. READOUT can be a simple permutation invariant function such as summation and mean or a more sophisticated graph-level pooling function. \cite{ying2018hierarchical, zhang2018end}

For our baseline, AGGREGATION is a simple function that summarizes the statistics of neighboring degree distribution by computing min, max, average, and standard deviation. We do not introduce trainable weights. The COMBINE function is a simple concatenation and READOUT is either histogram or empirical distribution operation. The number of iteration is $K=2$ in our case. From this point of view, \DOM{} captures the essential elements of GNN, and this may partially explain its effectiveness. However, it is quite surprising to us that without any learning we can still achieve results comparable to many GNNs. We suspect that the use of AGGREGATION or READOUT function may play an important role. Interestingly a recent paper\cite{xu2018powerful} explore the expressiveness of different pooling strategies and conclude that sum pooling is more powerful than mean and max pooling. 
To leverage this, we introduce  ``sum(DN($v$))'' as an extra node feature in the hope of achieving better results. We did a preliminary experiment on REDDIT 5K and REDDIT 12K but only achieved marginal improvement(0.5 percent). This might be due to the fact that our baseline does not involve any learning. Thus in the reported results, this ``sum(DN($v$))'' feature is not deployed. 
 
\paragraph{Discussion on  comparison to Weisfeiler-Lehman Kernel}
We can also see the similarity between \DOM{} and the Weisfeiler-Lehman kernel/isomorphism test with two iterations.  Both methods start from local node feature and build new feature from the previous step through graph topology. However, a key difference between WL kernel and \DOM{} is that the hashing step(label compression) of WL kernel {\bf do not} necessarily capture the local similarities: even two nodes with very similar neighborhood could be mapped to totally different labels, and perturbing the edges by a small amount will lead to completely different hashing features. \DOM{} instead uses the statistics of degree distribution of local neighborhood that is more robust and able to capture the node similarity. Empirically, we observe that it appears to strike a good balance between discriminating different local structures versus being robust to small differences.

Interestingly, in a recent paper \cite{xu2018powerful} that explores the discriminative power of graph neural networks showed that GNNs are at most as powerful as the WL test in distinguishing graph structures. By choosing right aggregation function, they develop graph isomorphism network(GIN) that can achieve the same discriminative power as WL test. However, in experiments, it is observed that GIN outperform WL kernel on social network dataset. One explanation the authors provide is that WL-tests are one-hot encodings and thus cannot capture the ``similarity'' between subtrees (while they can capture that whether they are ``the same'' or not). In contrast, a GNN satisfying certain criteria (see Theorem 3 of \cite{xu2018powerful}) generalizes the WL test by \textit{learning to embed} the subtrees to a feature space \emph{continuously}. This enables GNNs to not only discriminate different structures, but also learn to map similar graph structures to similar embeddings and capture dependencies between graph structures.   

We can see in the experiment section that after incorporating the features that able to capture similarity of graph structures, \DOM{} outperforms WL kernel by a large margin on social network dataset and achieves the result on par with GIN, even though no learning is involved in \DOM{}. 

\section{Experimental Results}

\paragraph{Datasets.} We collect commonly used graph datasets from existing graph classification literature as our benchmark data. 
For non-attribute graphs, we use the movie collaboration datasets IMDB BINARY and IMDB MULTI,  the scientific collaboration data COLLAB, the social network datasets REDDIT BINARY, REDDIT 5K, and REDDIT 12K. For non-attribute graphs, we use the protein dataset ENZYMES, PROTEINS and D\&D, chemical compounds datasets MUTAG, PTC, and NCI1. See appendix for a more detailed description of these datasets, including their statistics and properties. 

\paragraph{Experimental setup}
All experiments are performed on a single Intel Xeon CPU E5-2630 v4@ 2.20GHz $\times$ 40 and 64GB RAM machine. We compare our baseline with 6 state-of-the-art graph kernels: Weisfeiler-Lehman Kernel (WL)\cite{shervashidze2011weisfeiler}, Graphlet kernel (GK)\cite{shervashidze2009efficient}, Deep Graph Kernel (DGK)\cite{yanardag2015deep}, RetGK\cite{zhang2018retgk}, FGSD\cite{verma2017hunt}, Weisfeiler-Lehman optimal assignment kernel (WL-OA)\cite{kriege2016valid}, and 5 graph neural networks: PATCHYSAN (PSCN) \cite{niepert2016learning}, GRAPHSAGE\cite{hamilton2017inductive}, DIFFPOOL\cite{ying2018hierarchical}, Graph Isomorphism Network GIN\cite{xu2018powerful}, Diffusion-convolutional neural networks (DCNN) \cite{atwood2016diffusion}.
In particular, among above graph kernels, only FGSD and GK does not utilize label information for chemical/protein graphs, and for GNNs, the size of chemical/protein graphs are too small and their performance is not reported for attribute graphs in the most paper.
 All the result is taken from published paper except the baseline. The code for \DOM{} will is available on github.\footnote{\url{https://github.com/Chen-Cai-OSU/LDP}}.

\paragraph{Results.} 
The results are shown in Table 2-4.
For non-attribute graphs, our results show that combining our simple degree-based features and the kernel machine, we can beat WL, GK, DGK kernels by a large margarin consistently, and achieve the results on par with more recent graph kernels such as RetGK and WL-OA(no more worse than 2.5 percent) and perform even better on certain datasets. On average, we are slightly better than RetGK for non-attribute graphs. However, our method is much simpler and faster than all of the previous graph kernels.  What is more, we find that even using only linear kernel, which yielding our final model equivalent to local feature + linear SVM,  performance on REDDIT BINARY, REDDIT 5K, and REDDIT 12K rarely degrades. 

Comparing with GNN, although many accuracy data is not available, we can still clearly see that the state-of-the-art GNNs do not show significant improvement over our baseline. It is well known that neural networks are data hungry, so one possible explanation is that the size of the current dataset is limiting the representation power of GNNs.  Since our model can be seen as a simple variant of GNN where no end-to-end learning is involved, we interpret current result more as a dataset problem instead of an algorithm problem: current benchmark is no longer suitable for evaluating different algorithms.  A Large-scale ImageNet-like dataset for graphs is strongly needed for the evaluation of GNNs.

\begin{table}[]
\label{tab: t2}
\centering
\begin{tabular}{@{}llllllllllll@{}}
\toprule
                      & WL                & GK                & DGK      & RetGk     & FGSD       & WL-OA    & \DOM{}  &\DOM{}*\\ \midrule
COLLAB        & 74.8           & 72.8         & 73.1      & 81.0     & 80.0             & 80.7     & 78.1            &73.9 \\
IMDB BINARY   & 70.8         & 65.9         & 67.0      & 71.9     & 71.0               & -              & 75.4          & 67.7 \\
IMDB MULTI    & 49.8             & 43.9         & 44.6      & 47.7     & 45.2            & -              & 50.0            & 45.4 \\
REDDIT BINARY & 68.2         & 77.3         & 78.0     & 92.6     & 86.5            & 89.3        & 92.1          & 89.8 \\
REDDIT 5K     & 51.2              & 41.0         & 41.3      & 56.1     & 47.8           & -              & 55.9           & 54.2 \\
REDDIT 12K    & 32.6         & 31.8         & 32.2      & 48.7     & -                 & 44.4      & 47.8         & 46.7 \\ \midrule
Average        & 57.90        & 55.45        & 56.03    & 66.33    & -            & -         & 66.55        & 62.95\\ \bottomrule
\end{tabular}
\caption{Comparison with Graph Kernel. LDP * means using only linear SVM.}
\end{table}

\begin{table}[]
\label{tab: t3}
\centering
\begin{tabular}{@{}llllllllllll@{}}
\toprule
                       & PSCN & GRAPHSAGE & DIFFPOOL & GIN  & DCNN   & DGCNN    & \DOM{}   &\DOM{}*\\ \midrule
COLLAB                       & 72.6              & 68.25             & 75.50    &80.2    &52.1  & 73.7            & 78.1        &73.9\\
IMDB BINARY               & 71.0             & -                     & -            &75.1    & 49.1 & 70.0            & 75.4        & 67.7\\
IMDB MULTI                        & 45.2             & -                     & -            & 52.3    & 33.5  &     47.8            & 50.0        & 45.4\\
REDDIT BINARY           & 86.3            & -                    & -            &92.4    &-        &  -            &92.1         & 89.8\\
REDDIT 5K                   & 49.1             & -                     & -            &57.5    & -         &  -            &55.9         & 54.2\\
REDDIT 12K                 & 41.3             & 42.24             & 47.04     &-        &-         & -            & 47.8        & 46.7\\ \bottomrule
\end{tabular}
\caption{Comparison with graph neural networks. LDP * means using only linear SVM.}
\end{table}

For chemical graphs, we can treat them as graphs without the label, and apply the same method. We can also introduce one more feature for each node, which is simply the node label in the original data. Adding label information in this way is certainly not principled as our results are no longer invariant to label permutation.  However, our goal in this paper is not to handle attribute graphs so we settle on this hack. As we can see, \DOM{} is still quite good on MUTAG, PTC, and PROTEIN datasets that do not have rich label information. For graphs with rich labels such as ENZYME, DD, and NCI1(See details in appendix), \DOM{} and its variant is much worse than the other kernels that can better utilize label information. This also suggests that MUTAG, PTC, and PROTEIN are not sufficient to evaluate different methods.
\begin{table}[h]
\label{tab: t4}
\centering
\begin{tabular}{@{}llllllllllll@{}}
\toprule
                      & WL        & GK        & DGK       & PSCN      & RetGk     & FGSD         & WL-OA     & \DOM{}  & \DOM{} + Label \\ \midrule    
MUTAG       & 84.4 & 81.6     & 87.4     & 89.0       & 90.3       &  92.1        & 84.5         & 90.1          & 90.3 \\              
PTC         & 55.4 & 57.3         & 60.1     & 62.3    & 62.5       &  62.8        & 63.6         &  61.7           & 64.5     \\               
ENZYME    & 53.4 & -         & 53.4     & -         & 60.4      &  -             & 59.9         &  35.3         & 40.9     \\                
PROTEIN   & 71.2 & 71.7         & 75.7     & 75.0     & 75.8       &  73.4        & 76.4         &  72.7           & 73.7 \\         
DD           & 78.6  & 78.5         & -        & 76.2    & 81.6       &  77.1        & 79.2         &  75.5            & 77.1 \\              
NCI1          & 85.4  & 62.3       & 80.3    & 76.3     & 84.5        &  79.8        & 86.1         &  73.0        & 74.3 \\    \bottomrule
\end{tabular}
\caption{Comparison with other graph kernels for chemical graphs. \DOM{} + Label means adding label as an extra node feature.}

\end{table}

\textbf{Variants and limitations of baseline: }
Surprised by the performance of our baseline that is based on purely local node feature, one natural extension is to incorporate more sophisticated 1) node features and 2) edge features in the hope of capturing the local and global graph topology better and therefore improving the accuracy. 

To test this, we have experimented adding other node features(also more expansive to compute) such as closeness centrality, Fiedler vector(the second smallest eigenvectors of graph Laplacian), and Ricci curvature\cite{lin2011ricci} of graphs. Interestingly, we observe that none of the above features yields any significant improvement consistently across all datasets. 

For edge features, we compute all-pair-shortest path distance and add the histogram of distance distribution along with degree-based features for small chemical graphs(For the social networks dataset, it is too costly to compute all pair shortest path distance and we do not observe any improvement in the preliminary experiments). There is about 2 percentage improvement consistently over different datasets.   \footnote{We ignore MUTAG and PTC for the reason that the datasets are too small, and that even \DOM{} + Label can achieve the best result. } This indicates that our local method fails to capture more global information which is shown to be useful for chemical/protein classification
\footnote{For graphs, two vertex sets are called non-homometric if the multi-sets of distances determined by them are different. It is unknown whether there exists any distance metric under which two vertex sets of non-isomorphic graphs are always non-homometric; But it is easy to show that the shortest path distance does not satisfy the requirement:  a cycle of four vertices and a triangle with a pendant edge are non-isomorphic but have the same multi-set of all pairwise shortest path distances, i.e., \{1, 1, 1, 1, 2, 2\}}. 
\begin{table}[h]
\centering
\begin{tabular}{@{}ccccc@{}}
\toprule
                    & FGSD & GK   & \DOM{}  & \DOM{} + distance \\ \midrule
ENEZYME     & -           &-        & 35.3 & 37.2         \\
PROTEIN     & 73.4    & 71.7 & 72.7 & 74.7         \\
DD              & 77.1    & 78.5  & 75.5    & 77.5         \\
NCI1            & 79.8    &     62.3   & 73.0 & 75.6         \\ \bottomrule
\end{tabular}
\caption{Adding distance distribution improves the accuracy. Compared with FGSD and GK, which are the only two models that also does not use label information, we can see after adding all pair shortest path distance, we can match their result on all datasets except the NCI1 for FGSD. This indicates that using only local degree-based features is not enough for chemicals/protein classification}
\end{table}

\section{Conclusion and Discussion}
 
Most graph kernels aim to capture graph topology and graph similarity in the hope of improving classification accuracy. Our experiment suggests that this is not yet well-reflected on current benchmark datasets for non-attribute graphs. This calls for the construction of better and more comprehensive benchmark datasets for understanding and analyzing graph representations. On the other hand, we emphasize that in scenarios where both graph nodes and edges have certain associated features,  proper handling labels can significantly improve the classification accuracy, as is shown clearly for the NCI1 dataset. Graph kernels have been rather effective in incorporating node/edge attributes.  It will be an interesting question to see how to incorporate attributes in a more effective yet still simple manner in our graph representation. 
 
 Also, although there are large scale chemical graphs datasets available\cite{hachmann2011harvard, ruddigkeit2012enumeration}, a benchmark dataset that contains many large graphs is still missing. We plan to create such benchmark dataset for future use. In general, while not addressed in this paper, we note that understanding the power and limitation of various graph representations, as well as the types of datasets with respect to which they are most effective, are crucial, yet challenging and remain largely open. 

\section{Acknowledgement}
We thank Shuaicheng Chang and Siyuan Ma for insightful comments. We thank Han Fu for proofreading. 

\section{Datasets description}
\label{sec: dataset}
The statistics of the benchmark graph datasets used in the paper are reported in Table 6. We describe these datasets in detail in the next section.
\subsection{Non-attributed graph datasets}
\textbf{IMDB-BINARY}  \cite{yanardag2015deep} is a movie collaboration dataset that consists of the ego-networks of 1,000 actors/actresses who played roles in movies in IMDB. In each graph, nodes represent actors/actress, and there is an edge between them if they appear in the same movie. These graphs are derived from
the Action and Romance genres.

\textbf{IMDB-MULTI} \cite{yanardag2015deep} is generated in a similar way to IMDB-BINARY. The difference is that it is derived from three genres: Comedy, Romance, and Sci-Fi.

\textbf{REDDIT-BINARY}\cite{yanardag2015deep}  consists of graphs corresponding to online discussions on Reddit. In each graph, nodes represent users, and there is an edge between them if at least one of them respond to the other's comment. There are four popular subreddits, namely, IAmA, AskReddit, TrollXChromosomes, and atheism. IAmA and AskReddit are two question/answer based subreddits, and TrollXChromosomes and atheism are two discussion-based subreddits. A graph is labeled according to whether it belongs to a question/answer-based community or a discussion-based community.

\textbf{REDDIT-MULTI(5K)}\cite{yanardag2015deep} is generated in a similar way to REDDIT-BINARY. The difference is that there are five subreddits involved, namely, worldnews, videos, AdviceAnimals, aww, and mildlyinteresting. Graphs are labeled with their corresponding subreddits.

\textbf{REDDIT-MULTI(12K)} \cite{yanardag2015deep} is generated in a similar way to REDDIT-BINARY and REDDITMULTI(5K). The difference is that there are eleven subreddits involved, namely, AskReddit, AdviceAnimals, atheism, aww, IAmA, mildlyinteresting, Showerthoughts, videos, todayilearned, worldnews, and TrollXChromosomes. Still, graphs are labeled with their corresponding subreddits.

\subsection{Graphs with discrete attributes}

\textbf{MUTAG} \cite{debnath1991structure} consists of graph representations of 188 mutagenic aromatic and heteroaromatic nitro chemical compounds. These graphs are labeled according to whether or not they have a mutagenic effect on the Gram negative bacterium Salmonella typhimurium.

\textbf{PTC} \cite{helma2001predictive} consists of graph representations of chemical molecules. In each graph, nodes represent atoms, and edges represent chemical bonds. Graphs are labeled according to carcinogenicity on rodents, divided into male mice (MM), male rats (MR), female mice (FM), and female rats (FR).

\textbf{ENZYMES} and \textbf{PROTEINS} \cite{borgwardt2005protein} consist of graph representations of proteins. Nodes represent secondary structure elements (SSE), and there is an edge if they are neighbors along the amino acid sequence or one of three nearest neighbors in space. The discrete attributes are SSE types. The continuous attributes are the 3D length of the SSE. Graphs are labeled according to which EC top-level class they belong to.

\textbf{DD} \cite{dobson2003distinguishing} consists of graph representations of 1,178 proteins. In each graph, nodes represent amino acids, and there is an edge if they are less than six Angstroms apart. Graphs are labeled according to whether they are enzymes or not.

\textbf{NCI1} \cite{shervashidze2011weisfeiler} consists of graph representations of 4,110 chemical compounds screened for activity against non-small cell lung cancer and ovarian cancer cell lines, respectively.

\begin{table}[]
\centering
\begin{tabular}{@{}llllll@{}}
\toprule
Datasets      & graph \# & class \# & average\_nodes \# & average edges \# & label \# \\ \midrule
MUTAG         & 188      & 2        & 17.93             & 19.79            & 7        \\
PTC           & 344      & 2        & 14.29             & 14.69            & 19       \\
ENZYME        & 600      & 6        & 32.63             & 64.14            & 3        \\
PROTEIN       & 1113     & 2        & 39.06             & 72.82            & 3        \\
DD            & 1178     & 2        & 284.32            & 715.66           &  81        \\
NCI1          & 4110     & 2        & 29.87             & 32.30            & 37       \\
IMDB BINARY   & 1000     & 2        & 19.77             & 96.53            & -        \\
IMDB MULTI    & 1500     & 3        & 13.00             & 65.94            & -        \\
REDDIT BINARY & 2000     & 2        & 429.63            & 497.75           & -        \\
REDDIT 5K     & 4999     & 5        & 508.82            & 594.87           & -        \\
REDDIT 12K    & 12929    & 11       & 391.41            & 456.89           & -        \\ \bottomrule
\end{tabular}
\caption{Statistics of the benchmark graph datasets}
\end{table}

\newpage
\bibliography{ref}

\end{document}